\documentclass[10pt,twocolumn,letterpaper]{article}

\usepackage[pagenumbers]{cvpr} 

\usepackage{graphicx}
\usepackage{amsmath}
\usepackage{amssymb}
\usepackage{booktabs}
\usepackage{float} 
\usepackage{pifont}

\usepackage[pagebackref,breaklinks,colorlinks]{hyperref}

\usepackage[capitalize]{cleveref}
\crefname{section}{Sec.}{Secs.}
\Crefname{section}{Section}{Sections}
\Crefname{table}{Table}{Tables}
\crefname{table}{Tab.}{Tabs.}

\def\cvprPaperID{*****} 
\def\confName{CVPR}
\def\confYear{2022}

\begin{document}

\title{1st Place Solutions for UG$^2$+ Challenge 2022\\
ATMOSPHERIC TURBULENCE MITIGATION}

\author{
Zhuang Liu$^*$, Zhichao Zhao$^*$, Ye Yuan, Zhi Qiao, Jinfeng Bai, Zhilong Ji\\
Tomorrow Advancing Life (TAL) Education Group\\
{\tt\small \{liuzhuang7,zhaozhichao,yuanye8,qiaozhi1,baijinfeng1,jizhilong\}@tal.com}
}
\maketitle

\begin{abstract}
In this technical report, we briefly introduce the solution of our team ``summer'' for Atomospheric Turbulence Mitigation in UG$^2$+ Challenge in CVPR 2022. In this task, we propose a unified end-to-end framework to reconstruct a high quality image from distorted frames, which is mainly consists of a Restormer-based image reconstruction module and a NIMA-based image quality assessment module. Our framework is efficient and generic, which is adapted to 
both hot-air image and text pattern. Moreover, we elaborately synthesize more than 10 thousands of images to simulate atmospheric turbulence. And these images improve the robustness of the model. Finally, we achieve the average accuracy of 98.53\% on the reconstruction result of the text patterns, ranking 1st on the final leaderboard.
\end{abstract}
\footnotetext[1]{Authors contribute equally.}

\section{Introduction}
\label{sec:intro}
The Atomospheric Turbulence Mitigation in UG$^2$+ Challenge in CVPR 2022 is a part of the Workshop on UG$^2$+ Prize Challenge. The task aims to promote the development of new image reconstruction algorithms for imaging through atmospheric turbulence.

\begin{figure}[t] 
    \centering
    \includegraphics[width=0.4\textwidth]{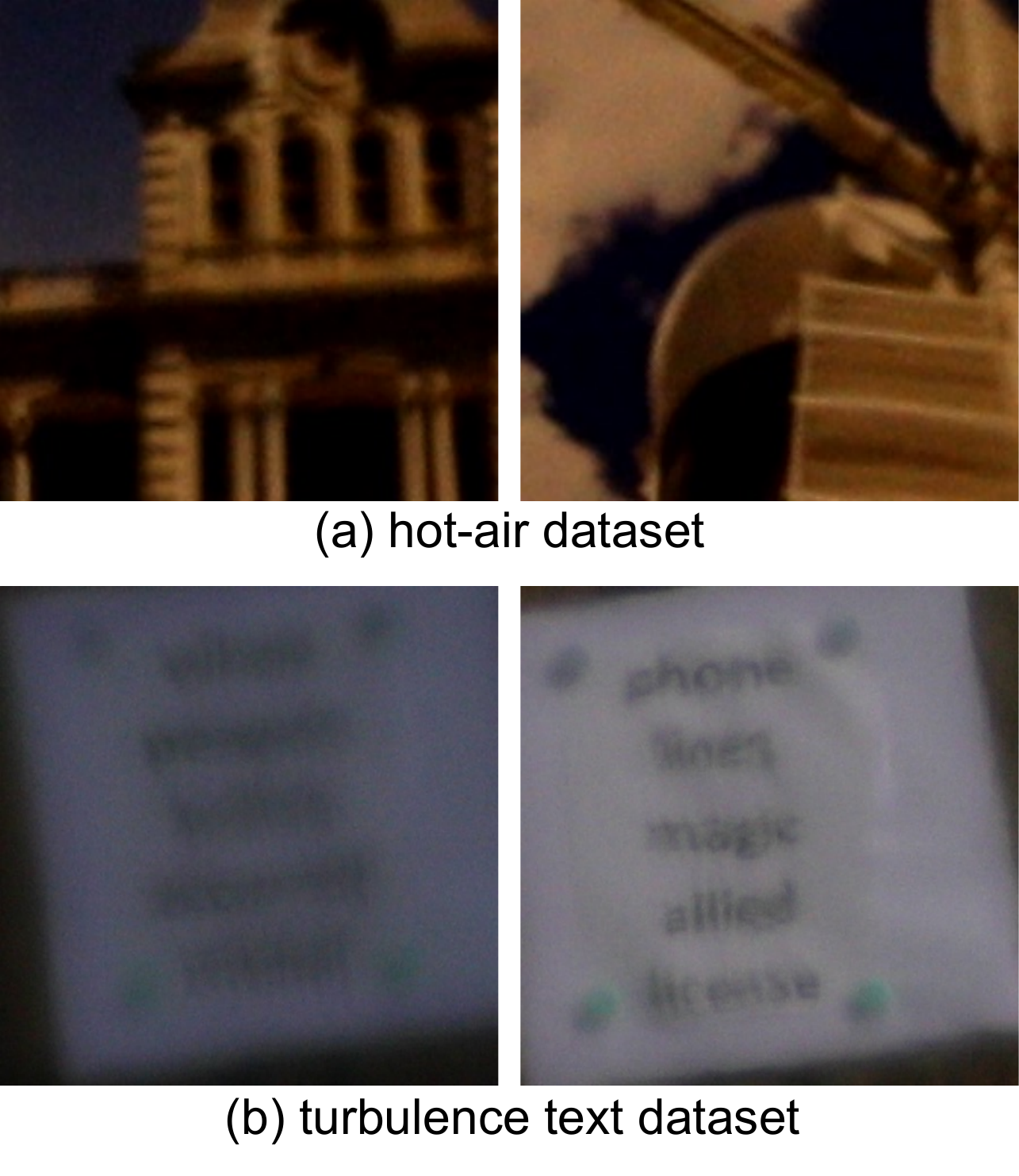}
    \caption{Sample images from (a) hot-air dataset and (b) turbulence text dataset} 
    \label{sample images}
\end{figure}

The training data is generated \emph{via} official provided atmospheric turbulence simulator. The proposed framework is evaluated with hot-air dataset (Fig. \ref{sample images}(a)) and turbulence text dataset (Fig. \ref{sample images}(b)). Hot-air dataset contains 50 sequences and text dataset contains 500 sequences. Each sequence contains 100 turbulence distorted frames. A high quality image could be reconstructed from all or part of 100 frames. Peak signal-to-noise ratio (PSNR) is the metric for evaluating the qualities of recovered hot-air images. The reconstruction result of the text images is based on the average accuracy of three existing scene text recognition algorithms (CRNN \cite{shi2016end}, ASTER \cite{shi2018aster} and DAN \cite{wang2020decoupled}).


\section{Overview of Methods}

\begin{figure*}[t]
  \centering
  \includegraphics[width=0.90\linewidth]{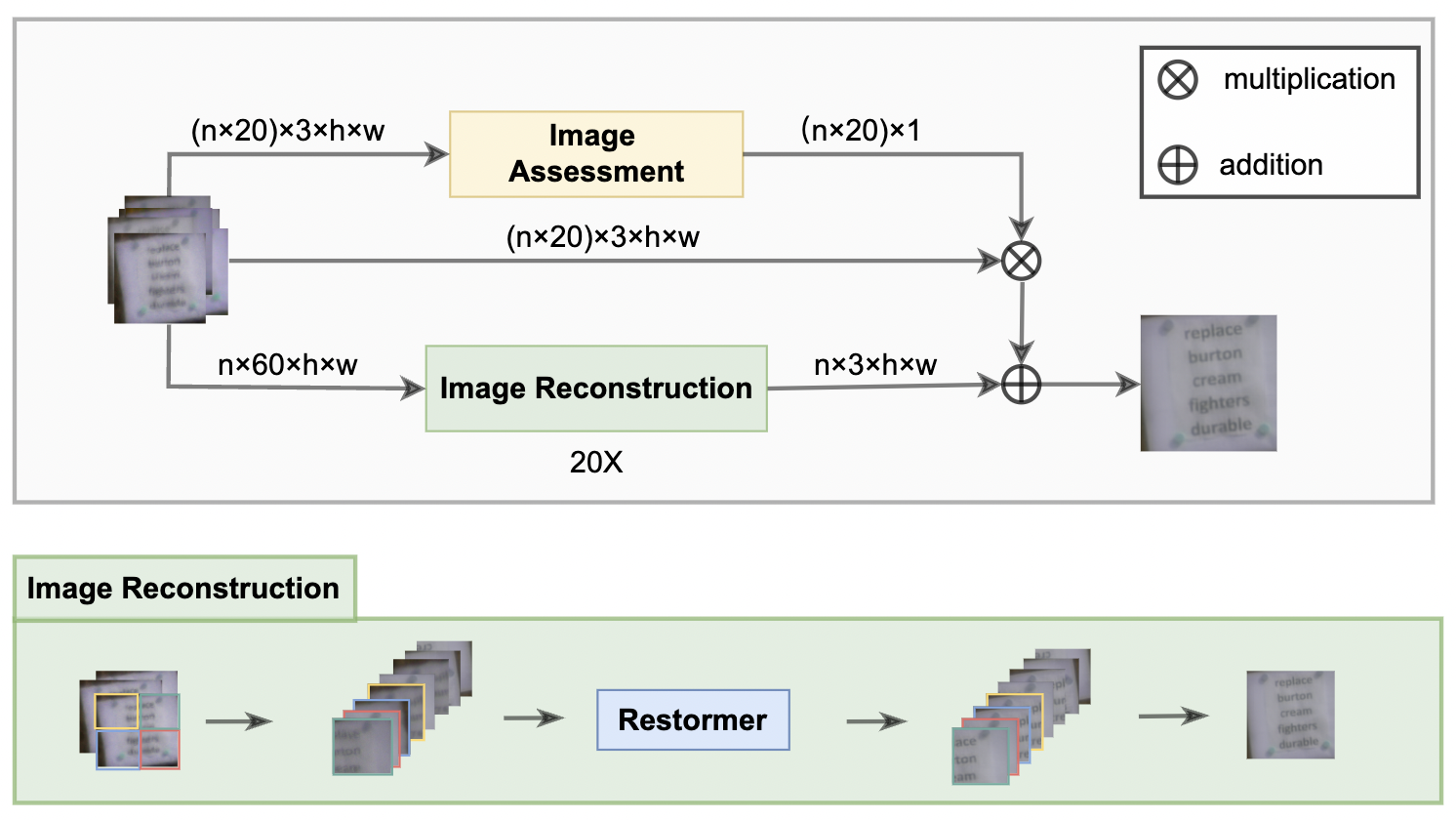}
   \caption{Structure of proposed framework, which consists of a image assessment module and a image reconstruction module.}
   \label{fig:model}
\end{figure*}

In this section, we will introduce our method used in this challenge. As shown in Fig. \ref{fig:model}, our unified framework comprises an image reconstruction module and an assessment module. Given degraded images $\mathcal{X}$, the image reconstruction module is used to get the refined features $\mathcal{F^{'}}$. The image assessment module is applied to the degraded image $\mathcal{X}$ to generate residual image $\mathcal{R}$ to which refined features is added to obtain the restored image: $\mathcal{F} = \mathcal{F^{'}} + \mathcal{R}$.


\subsection{Image Reconstruction Module}

Following recent studies on image restoration, we adopt Restormer \cite{Zamir2021Restormer} as the backbone. The Restormer adopts a 4-level symmetric encoder-decoder and each level of encoder-decoder contains
multiple Transformer blocks are increased from the top 4 blocks to bottom 8 blocks. To reduce the computational complexity, we apply sliding window (size of 128 x 128) with stride equals 100 to divide the input images into small patches and the model is trained on small patches.

In this task, each sequence contains 100 turbulence distorted frames. To fully use the information from different frames, 20 random selected frames are given to restore a high quality image. As a result, the Restormer takes 20 image patches X, with
size of (20 x 3) × H × W, where H and W are both equal to 128.


\subsection{Image Assessment Module}

Drawing on the residual design, a image quality assessment module$^\dagger$ is incorporated in the framework. This module relies on a non-reference image quality assessment method, named Neural Image Assessment (NIMA) \cite{talebi2018nima}. Given the degraded images, image assessment module outputs quality scores. Degraded images are multiplied by the quality scores and added to output of image reconstruction module to get the final restoration images.



\subsection{Loss Function}

The overall loss function consists of two parts and is defined as follows:
\begin{equation}
\displaystyle{\mathcal{L}=\alpha\mathcal{L}_{L1}+ (1-\alpha)\mathcal{L}_{S S I M}}
\end{equation}
where $\mathcal{L}_{L1}$ is a $L1$ regression loss and $\mathcal{L}_{SSIM}$ is a $SSIM$ regression loss. $\alpha$ is a hyper-parameter to balance the L1 loss and SSIM loss, which is set to 0.5 during the training.


\begin{figure}[t] 
    \centering
    \includegraphics[width=0.45\textwidth]{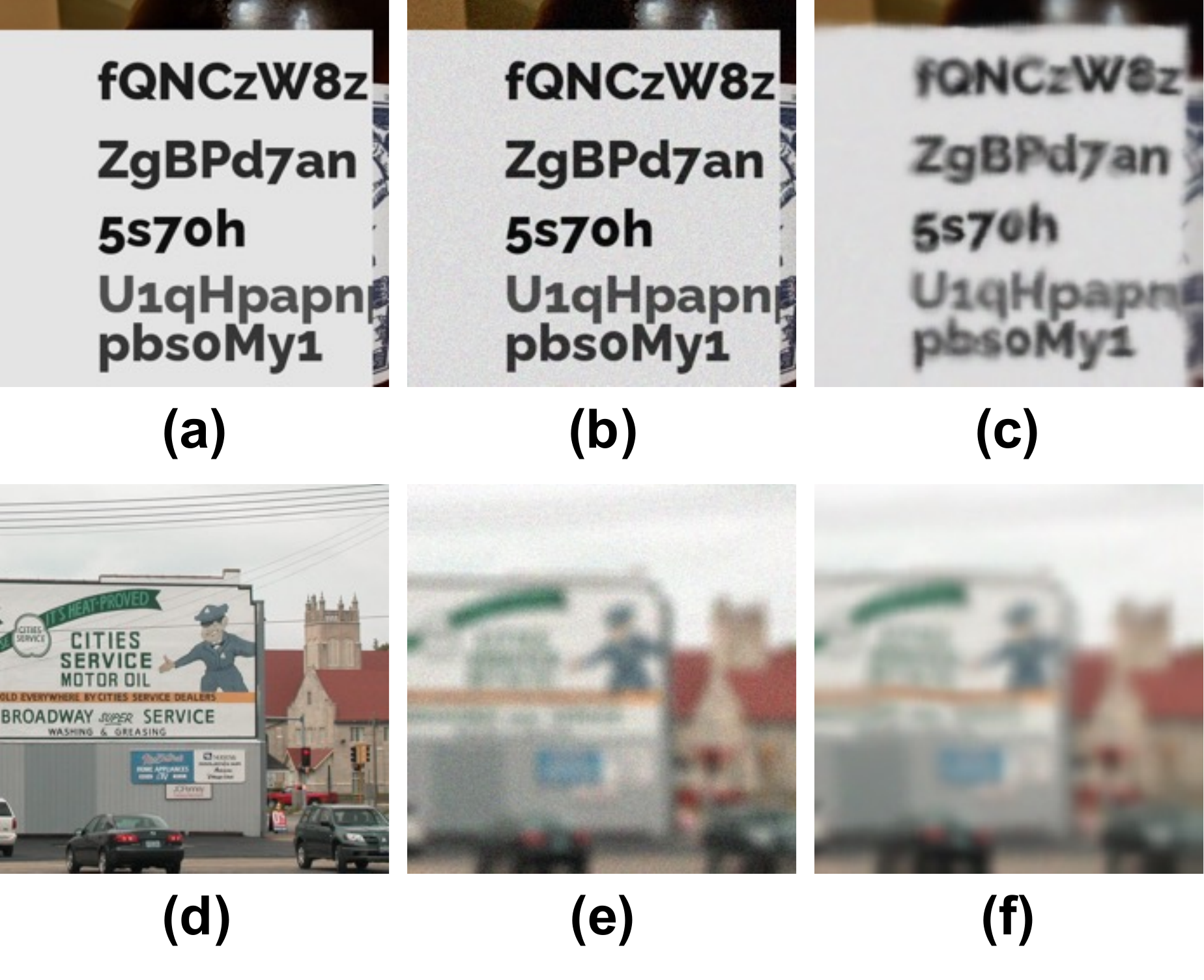}
    \caption{(a) Synthesized image with clean background. (b) Degraded image with Gaussian noise, etc. (c) Generated turbulence image. (d) Scene text image. (e) Degraded image with Gaussian noise, etc. (f) Generated turbulence image.}
    \label{data generation}
\end{figure}

\begin{table*}[htbp]
    \setlength\tabcolsep{3pt}
    \centering
    \caption{Ablation studies on dry-run's turbulence text dataset. The effect of recognition performance with regard to the five basic settings: prtraining, finetuning, small patches, rotation and multiple inference. (\ding{51}) denotes the setting existence while (\ding{53}) indicates the setting absence. Average accuracy is reported as a percentage (\%)}
    \label{ablation studies}
    \begin{tabular}{c|ccccc|c}
    \toprule[1.5pt]
    \textbf{Model} & \textbf{Pretaining} & \textbf{Fine-tuning} & \textbf{Small Patches} & \textbf{Rotation} & \textbf{Multiple Inferences} & \textbf{Average Accuracy} \\ \midrule[1pt]
    U-Net(ResNet-18) & \ding{51}&  \ding{53} &  \ding{53} &  \ding{53} &  \ding{53} & 35.0 \\
    Restormer & \ding{51}&  \ding{53} &  \ding{53} &  \ding{53} &  \ding{53} & 53.3 \\
    Restormer & \ding{51}&  \ding{51} &  \ding{51} &  \ding{53} &  \ding{53} & 84.6 \\
    Restormer & \ding{51}&  \ding{51} &  \ding{53} &  \ding{51} &  \ding{53} & 73.3 \\
    Restormer & \ding{51}&  \ding{51} &  \ding{51} &  \ding{51} &  \ding{53} & 86.0 \\
    \textbf{Restormer} & \textbf{\ding{51}}&  \textbf{\ding{51}} &  \textbf{\ding{51}} &  \textbf{\ding{51}} &  \textbf{\ding{51}} & \textbf{93.3} \\

    \bottomrule[1.5pt]
    \end{tabular}
\end{table*}

\section{Data Preparation}

To close the domain of turbulence text images and hot-air images, we synthesize over 8000 text images (Fig. \ref{data generation}(a)) with clean backgrounds and randomly select over 2000 scene text images (Fig. \ref{data generation}(d)) from CTW\cite{yuan2019ctw}, COCO Text\cite{coco} and DIV2K\cite{div2k} respectively. The text generation procedures can be summarised as follows. After acquiring suitable word instance and background image, a text sample is rendered using a randomly selected font and size and transformed according to the random orientation. Then, we add extra Gaussian noise, Gaussian blur, brightness, etc. to further degrade these images (Fig. \ref{data generation}(b, e)). Finally, the official provided atmospheric turbulence simulator is applied to further adding turbulence (Fig. \ref{data generation}(c, f)). The setting of aperture size (D), Distance (L), Refractive Index Structure Constant ($C{_n^2}$) and Spatial Correlation (corr) to the simulator are shown as follow:

$\begin{cases}
\textbf{D}   \in [0.06143, 0.091254]  \\
\textbf{L}   \in [200, 400] \\
{\bf C{_n^2}}   \in [5.7386e-14, 9.7386e-14] \\
\textbf{corr}  \in [-1, 0] \\
\end{cases}$

We synthesized a total of 8706 images with clean backgrounds and 2569 images with street view backgrounds. We randomly selected 7706 images with clean backgrounds and 1569 images with street view backgrounds for pre-training. The last 2000 images are used for fine-tuning.

\section{Experiments}


\subsection{Implementation Details}

The Restormer is implemented in PyTorch. We use eight Nvidia Tesla
V100 with 32GB RAM to train our model with batch size 32. The Adam
optimizer \cite{kingma2014adam} is used during the training process, in which weight decay is set at 1e-4. The learning rate starts 1e-4 decays to 1e-10 following the cosine schedules. For pre-training, the model is optimized for 108 epochs. For fine-tuning, the model is optimized for 140 epochs. The common augments such as rotation, random cropping, etc. are also adopted to improve the robustness of the model.






\subsection{Bag of Tricks for Inference}
\label{Inference}

\subsubsection{Using Small Patches}

To reduce the computational complexity, small patches are used to optimize the model. During the inference, we also employ this process for consistency. Degraded small patches are first to be reconstructed. Then, integrate these high quality patches together to get the final image.


\subsubsection{Rotation}

Using degrade images with different rotation angles is proved to be an effective trick to get better restoration images. The degraded images randomly rotate at 0, 90, 180, or 270 degrees before input the model. After the reconstruction, the rotated patches will be firstly recovered from the rotation and then integrated to get the final images. The effects of rotation treatment will be discussed in the ablation studies section.


\begin{figure}[t]
  \centering
  \includegraphics[width=0.99\linewidth]{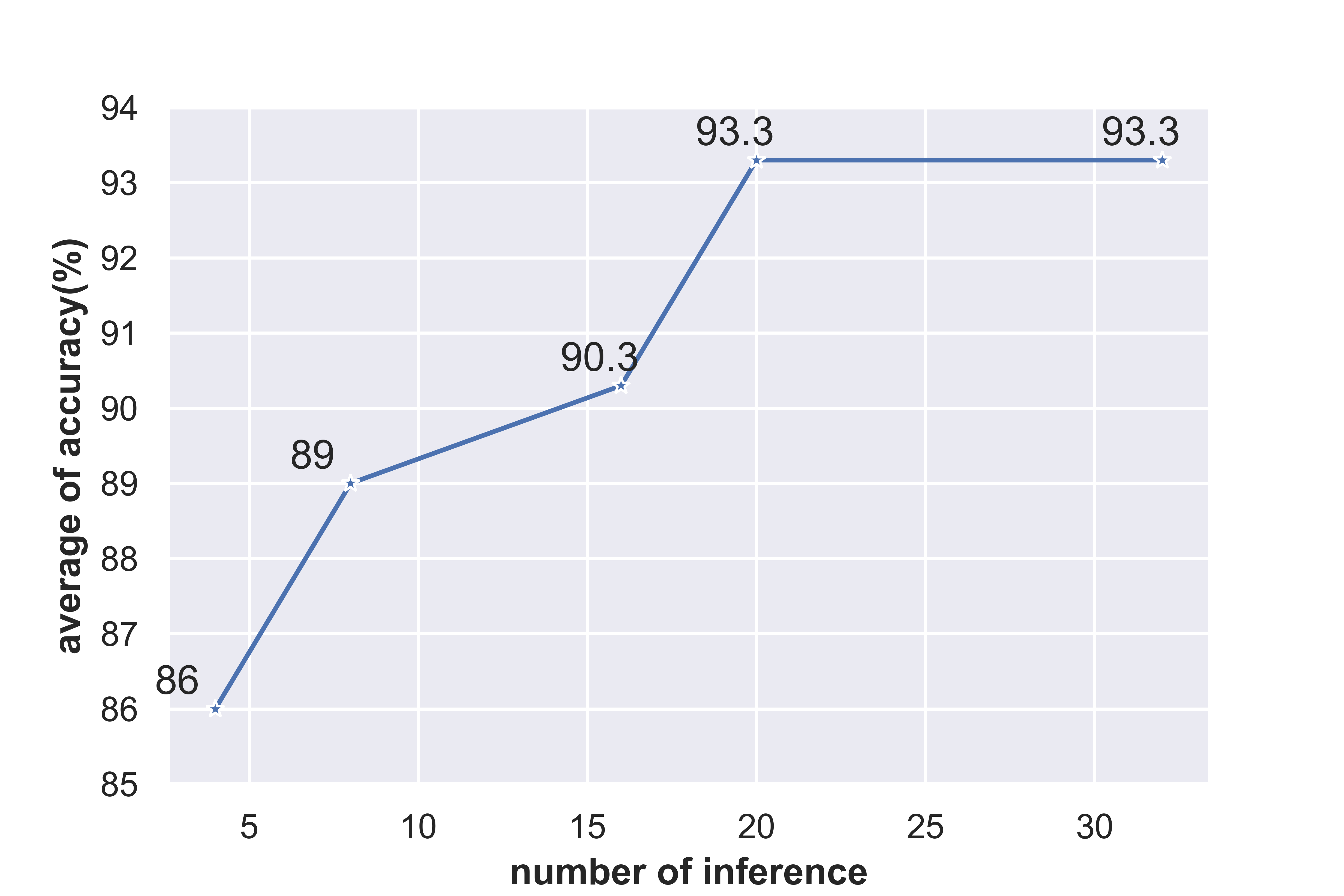}
   \caption{Average accuracy of reconstructed turbulence text images as a function of number of inference}
   \label{fig:repeat}
\end{figure}

\subsubsection{Multiple Inferences}

As the degradation of 100 frames for each sequence is different. In order to avoid missing the key frames, we repeat the inference procedure for 20 times with different degraded frames. And the average of these 20 results is used as the final result. Fig. \ref{fig:repeat} shows the shows the average accuracy as a function of the number of inferences. It is clear to notice that as the number of inferences increases, the average accuracy also increases gradually.








\subsection{Ablation Studies}

In this subsection, we perform ablation studies to analyze the impact of prtraining, finetuning, small patches, rotation and multiple inference. We trained all models from scratch and evaluated their performance on turbulence text dataset. The dry-run results are summarized in Table \ref{ablation studies}.

As it is illustrated in Table \ref{ablation studies}, U-Net is the default model, while using Restormer as the backbone improves the average accuracy by 18.3\%. Fine-tuning with small patches and Fine-tuning with rotation can further increase the average accuracy to 84.6\% and 73.3\%, respectively. Combined with these three tricks, the average accuracy rises to 86.0\%. Multiple inference also is effective, which promote the average accuracy to 93.3\% as the final result.



\begin{figure}[h]
  \centering
  \includegraphics[width=0.99\linewidth]{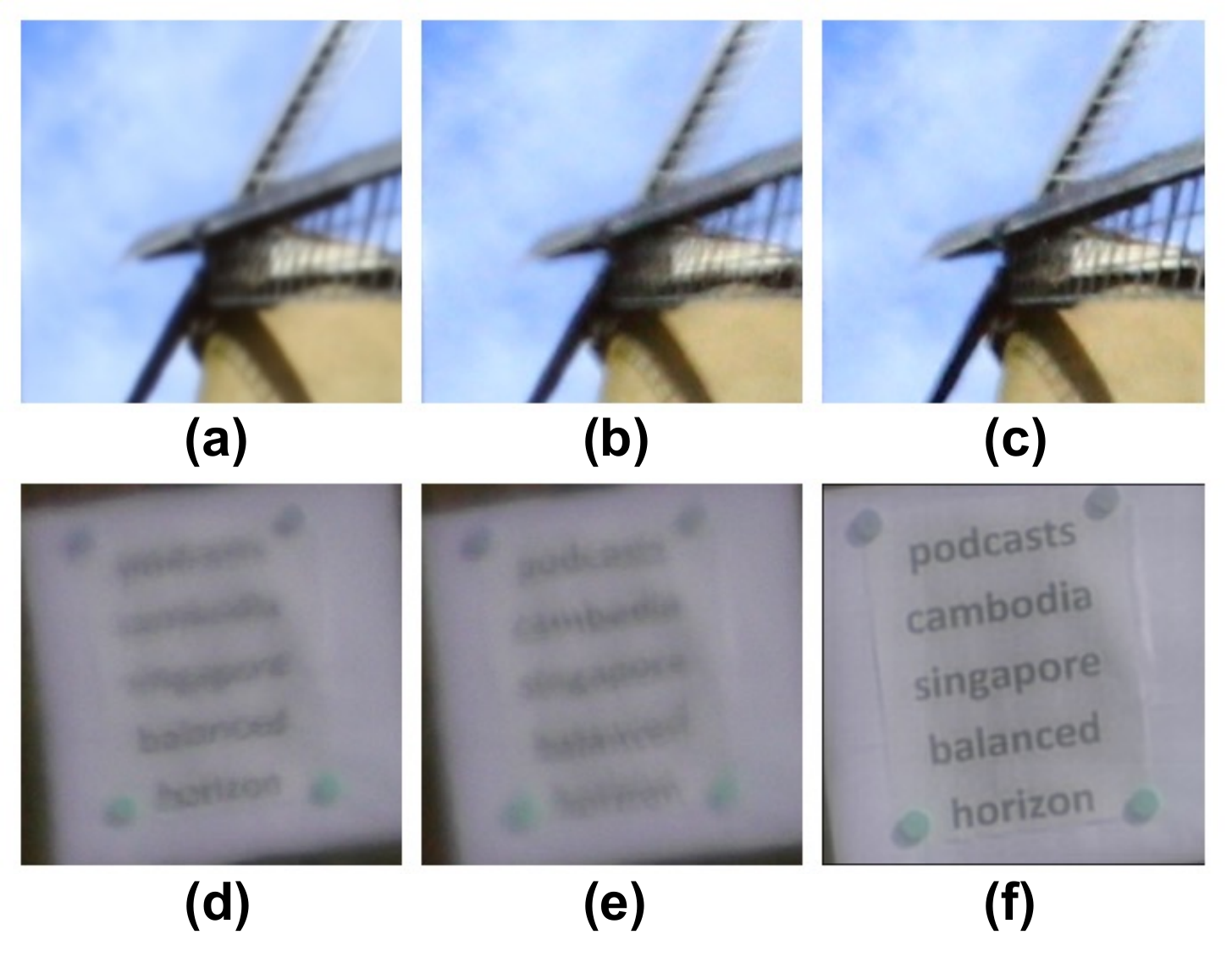}
   \caption{Samples of reconstruction results. (a) and (b) are from hot-air dataset. (d) and (e) are from turbulence text dataset. (c) and (f) are restored images.}
   \label{fig:image}
\end{figure}

\subsection{Visualization}

In this section, we choose some examples to visualize the restoration effect. As shown in Fig. \ref{fig:image}, the edge and contours of hot-air images are more clearer after restoration. For turbulence text images, the texts are much clearer.

\section{Conclusion}
In our submission to the track 3 in UG$^2$+ Challenge in CVPR 2022, we propose an  efficient and generic end-to-end framework to reconstruct a high quality image from atmospheric turbulence image. We adopt Restormer as the backbone and combine a image quality assessment module to further improve the performance. Besides, we elaborately synthesize images with both clean backgrounds and street view backgrounds to close the domain of hot-air images and turbulence text images. Finally, we achieve the 1st place on the final leaderboard. In the future, We will explore more efficient method to process this task.

%

{\small
\bibliographystyle{ieee_fullname}
\bibliography{egbib}
}

\end{document}